# Spatially-aware station based car-sharing demand prediction

Dominik J. Mühlematter[a,*], Nina Wiedemann[a], Yanan Xin[a] and Martin Raubal[a]

[a] Chair of Geoinformation Engineering, ETH Zürich, Zürich, Switzerland

ABSTRACT

In recent years, car-sharing services have emerged as viable alternatives to private individual mobility, promising more sustainable and resource-efficient, but still comfortable transportation. Research on short-term prediction and optimization methods has improved operations and fleet control of car-sharing services; however, long-term projections and spatial analysis are sparse in the literature. We propose to analyze the average monthly demand in a station-based car-sharing service with spatially-aware learning algorithms that offer high predictive performance as well as interpretability. Our study utilizes a rich set of socio-demographic, location-based (e.g., POIs), and car-sharing-specific features as input, extracted from a large proprietary car-sharing dataset and publicly available datasets. We first compare the performance of different modeling approaches and find that a global Random Forest with geo-coordinates as part of input features achieves the highest predictive performance with an R-squared score of 0.87 on test data. While a local linear model, Geographically Weighted Regression, performs almost on par in terms of out-of-sample prediction accuracy. We further leverage the models to identify spatial and socio-demographic drivers of car-sharing demand. An analysis of the Random Forest via SHAP values, as well as the coefficients of GWR and MGWR models, reveals that besides population density and the car-sharing supply, other spatial features such as surrounding POIs play a major role. In addition, MGWR yields exciting insights into the multiscale heterogeneous spatial distributions of factors influencing car-sharing behaviour. Together, our study offers insights for selecting effective and interpretable methods for diagnosing and planning the placement of car-sharing stations.

## 1. Introduction

In light of the necessity of transforming the transportation sector to reduce $CO_2$ emissions, car-sharing services have great potential to supplement or complement available transport options. In contrast to controversies around micromobility such as e-scooters, car-sharing was shown to effectively reduce car ownership [Mishra et al., 2015, Martin and Shaheen, 2011, Liao et al., 2020]. Additionally, it could reduce parking spaces and increase resource efficiency in urban environments [Glotz-Richter, 2016]. Despite these societal and environmental benefits brought by car-sharing, it has been challenging for car-sharing services to be profitable due to high investments and suboptimal utilization. Research has thus aimed to support the development of sustainable and profitable services with operations research and data-driven methodology for short-term demand prediction [Illgen and Höck, 2019, Wang et al., 2021, Zhu et al., 2019].

In this context, it has been shown that spatial and socio-demographic factors play an important role in car-sharing membership and utilization patterns [Amirnazmiafshar and Diana, 2022, Becker et al., 2017, Juschten et al., 2019]. For example, it was found that car-sharing is most popular in urban areas, among young people who are predominantly male [Becker et al., 2017], and among people who are less "car-oriented" and have a higher rate of public transport subscriptions. Such insights can not only help car-sharing companies attract new customers, but can also be incorporated into computational methods for design and operational decisions. For example, socio-demographic features were incorporated in short-term predictive models such as the work by Cocca et al. [2020], or for simulation [Ciari et al., 2016]. However, socio-demographic and spatial features often exhibit spatial dependence and spatial heterogeneity effects that might not be sufficiently captured by *spatially-implicit* prediction models. This issue may be negligible in short-term demand predictions (usually with a prediction horizon of hours or days) where the strong temporal dependency often plays a more important role in the prediction. However, it becomes crucial in long-term demand estimation, where we define "long-term demand" as the average demand over weeks or months. In contrast to short-term demand estimation which is relevant for daily operational decisions, long-term estimates are necessary for system planning and design, for example, to assess the potential demand at a new car-sharing station. Planning new stations or other design choices requires high investments and strongly confined operations in the daily business. It is therefore crucial to gain a better understanding of the factors that are relevant for long-term decision-making. There have been very limited studies to predict long-term car-sharing demand and, in particular, to foster the interpretability of prediction models. So far, studies either aim at improving the accuracy of short-term prediction with black-box models, or at explaining the influences of predictors on long-term demand with simple linear regression models, ignoring non-linear effects or spatial heterogeneity.

We aim to close this gap with interpretable and spatially explicit models for inferring car-sharing demand using socio-demographics and spatial features. Specifically, in our study, we predict the average monthly demand on a station level, with the goal of providing an explainable tool to assist in planning new stations. Besides a rich set of population features extracted from census data, we include spatial characteristics such as points-of-interest (POI) density and public transport accessibility in the models. Our study is conducted on a national-scale dataset by a car-sharing vendor in Switzerland comprising the demand at 1641 stations. Going beyond a simple regression analysis, we leverage SHAP values [Lundberg and Lee, 2017] to understand non-linear relations learned via the Random Forest model, and analyze the spatial heterogeneity of predictors via the Multi-Scale and the standard Geographically Weighted Regression models. For validation, these models are compared to other local and global models in terms of predictive and analytical powers. Our results demonstrate the strong ability of the RF model with geocoordinates and GWR model in predicting the long-term demand at new stations, with an R-squared score of up to 0.87 on a left-out test set of stations. Our study provides a critical discussion of the interpretability and effectiveness of global and local methods for demand prediction, and can guide further efforts in related applications.

*Corresponding author
dmuehlema@ethz.ch (D.J. Mühlematter)
ORCID(s): 0000-0001-6800-9114 (D.J. Mühlematter)

In the following, we will first discuss related work in section 2, then explain the data processing (section 3) and methodology (section 4), present and discuss our results in sections 5 and 6 respectively, and finally conclude our paper in section 7.

## 2. Related work

### 2.1. Car-sharing system optimization and demand prediction

Car-sharing as a relatively new transportation concept has received much attention in recent literature [Shaheen and Cohen, 2013]. Ferrero et al. [2018] for example review 137 papers on car-sharing services and summarize business perspectives, user group analysis, and operational designs developed in the literature. Our study lies between system design analysis and demand forecasting, since we aim to predict long-term demand at new stations with learning methods. Most machine learning approaches are, however, built for short-term demand prediction. For example, Wang et al. [2021] estimate the demand with a neural network for the purpose of car relocation, and Zhu et al. [2019] and Yu et al. [2020] apply Graph Convolutional Networks or LSTMs respectively to improve the short-term predictive performance with more complex models. Further work on short-term demand prediction can be found in the literature on *bike* sharing, e.g. by Xu et al. [2018] and Qiao et al. [2021]. While these models are powerful for short-term forecasts based on the demand history because they can efficiently model temporal patterns and detailed interactions between stations, other methods and especially other input features are needed for long-term demand estimation. Specifically, short-term prediction requires time series analysis methods to capture how demand changes with the time of the day, with previous demand, and with external factors. Long-term prediction is mainly driven by external factors such as the location and socio-demographic contexts. Importantly, spatial analysis becomes crucial in this problem setting. Cheng et al. [2022], for example, estimate station-based long-term car-sharing demand using a large variety of features as input, ranging from social network data to land use. They analyze spatial autocorrelations extensively and improve the understanding of how spatial factors influence car-sharing demand. On the other hand, system design choices are sometimes tackled with simulations instead of predictive methods. For example, Ayed et al. [2015], Balac et al. [2015], Ciari et al. [2013] use the MATSim simulator for simulating return- or one-way car-sharing systems (see [Ciari et al., 2016] for a review). However, such simulations are complex and rely on many assumptions about the population, their travel decision, and other transportation options.

### 2.2. Influence of socio-demographics on car-sharing

Amirnazmiafshar and Diana [2022] review the effect of socio-demographic factors on car-sharing interest, comparing different car-sharing designs such as free-floating vs. station-based. They analyze the effect of age, demographic region, and travel behavior, and find that free-floating services attract comparably more young male and high-income users. Since the *Mobility* car-sharing service in Switzerland is one of the oldest and largest car-sharing services worldwide, studies have specifically analyzed its user groups and usage patterns [Becker et al., 2017, Juschten et al., 2019]. Few works have used socio-demographic features for demand prediction, but they barely improve the short-term predictability of demand [Cocca et al., 2020]. The work by Kumar and Bierlaire [2012] may be most related to our work since they estimate long-term demand with OLS regression, and utilize the model to optimize the layout of new stations.

### 2.3. Spatial regression methods

Aside from socio-demographic features, the demand of a car-sharing station is also affected by spatial features, such as the distance from public transport stations or other POIs, or land use in the surrounding area. We hypothesize that both socio-demographic and spatial factors can explain long-term demand, but are subject to spatial heterogeneity, which cannot be captured sufficiently by global models [Wiedemann et al., 2023]. We therefore apply Geographically Weighted Regression (GWR) [Brunsdon et al., 1998] which was developed to acknowledge potential differences in the influence of factors across space. In GWR, the model is fitted on local neighborhoods of points, thereby allowing the coefficients to vary in space. Many variants of this basic idea exist [LeSage, 2004], and Comber et al. [2022] provide a route map to decide when to apply standard, mixed [Brunsdon et al., 1999], or multi-scale [Fotheringham et al., 2017] GWR. Later, the idea of GWR was combined with the predictive power of Random Forest (RF) models in the form of Geographical Random Forests (GRF) [Georganos et al., 2021].

These local methods were successfully applied for the *analysis* and *prediction* of various spatial phenomena, including transportation or mobility research. For example,

Nowrouzian and Srinivasan [2013] relate travelled distance to land use factors with GWR, and Wei et al. [2021] explain spatial differences in shared bicycle usage. Hosseinzadeh et al. [2021] analyze shared e-scooter trip generation with GWR, and Cardozo et al. [2012] predict transit ridership with spatial models. An example of long-term demand prediction for a shared system can be found in the study by Guidon et al. [2020], where spatial models were applied to estimate the potential demand for a bike-sharing system in a new city. They showed that spatial models outperform basic (spatially-implicit) Random Forest models. Other studies simply analyze the influence of spatial features on car-sharing demand. For example, Stillwater et al. [2009] explain car-sharing demand with features of the built environment via OLS-regression. Wagner et al. [2015] conduct a study closely related to ours where they relate the demand for shared vehicles (free-floating) to surrounding POI data with a GWR model.

In this study, we extend previous works by investigating the utility of various spatially-explicit models (inc. linear and non-linear models) in analyzing and predicting car-sharing demand. We compare these models' strengths and weaknesses in terms of interpretability, predictive power, and complexity of implementation, and offer guidelines for selecting the most proper models for different applications. We also provide a case study to illustrate the effectiveness of RF (with geocoordinates) and GWR in the application of planning new car-sharing stations.

## 3. Data and preprocessing

### 3.1. Data sources

This study pools several public and proprietary data sources, including car-sharing behavior, socio-demographic population statistics, and mobility behavior. Our primary source is a dataset provided by Mobility Cooperative, a car-sharing system in Switzerland that is one of the biggest in Europe. In 2022, they had over 3120 cars operating at 1530 supply stations [Cooperative, 2022], spreading across Switzerland and covering urban as well as rural areas. The dataset contains information about the locations, bookings, and vehicles of the car-sharing stations from the period of 1st of January 2019 until 26th of February 2020. Since the number of contexts and user-related features in the car-sharing service dataset is limited, we enriched the data with socio-demographic and geographic information details on a station level. We utilize census studies from the Federal Statistical Office of Switzerland [Federal Office for Spatial Development (ARE), Federal Statistical Office (FSO), 2017], specifically the Mobility and Transport Microcensus, the population, household and business statistics, and public transportation information. First, the Mobility and Transport Microcensus is a statistical survey on travel behaviour in Switzerland in 2015, published by the Federal Office for Spatial Development. More than 57'000 households participated. The dataset contains personal features such as the number of cars or the income per household. Further, the dataset provides location information and geocoordinates of the households. Secondly, the Federal Office for Spatial Development created a spatial classification of the public transport accessibility for all areas in Switzerland based on the criteria of transport mode, frequency, and distance to the nearest public transport station (ÖV-Güteklassen) [Federal Office of Spatial Development (ARE), 2022]. The classification consists of an ordinal scale that measures the performance of the public transport system.

Third, the population and household statistic (STATPOP 2020) is a dataset from the Statistical Office of Switzerland [Federal Statistical Office, 2021b]. Socio-demographic features like the population density, the average household size, or the average age of the inhabitants are available in a spatial resolution of 100 m. Similarly, the structural business statistics (STATENT 2019) provides comprehensive information on the structure of the Swiss economy [Federal Statistical Office, 2021a]. Examples of such information are the number of jobs or the number of workplaces per economic sector. The Statistical Office of Switzerland publishes this dataset using a hectare resolution.

Finally, we extracted Points of Interest (POI) from OpenStreetMap in several thematic categories (public, health, leisure, food, and accommodation) [OpenStreetMap contributors, 2022]. The POIs were selected based on our experience and expert knowledge.

### 3.2. Data fusion and preprocessing

The data from the Mobility car-sharing Service was enriched with features from the previously mentioned data sources. Figure 1 illustrates the procedure.

First, we filtered and cleaned the car-sharing dataset. Initially, the dataset comprises 1641 car-sharing stations with 1'471'125 reservations during the covered time period. Stations without or with wrong coordinates[1] as well as duplicates of stations were removed. Further, trips with unrealistic durations (> 500 hours) and distances (< 0 km, > 500 km) were not considered. Since over 99.3% of all car-sharing trips were return trips, only these were taken into account. In the resulting filtered dataset, 1439 stations with 1'248'342 bookings remain. The distribution of the number of reservations per station is shown in Figure 3a ("Ground truth").

In the next step, we calculated Voronoi Cells for each station using the Python library GeoVoronoi [Konrad, 2022]. A Voronoi cell for a specific station covers the area with the shortest distance to the respective station. Then, we counted the points of interest by category per Voronoi cell, and calculated the density by dividing this value by the area of the cell as shown in Figure 2b.

The location information (e.g., distance from a household to the nearest hospital) from the Microcensus dataset was fused with a spatial nearest neighbour join: Each household was assigned to the nearest car-sharing station as shown in Figure 2a. The average value per feature from all households belonging to the same station was calculated.

[1] Those stations were usually labelled as "internal use" of the car-sharing system and can therefore safely be removed

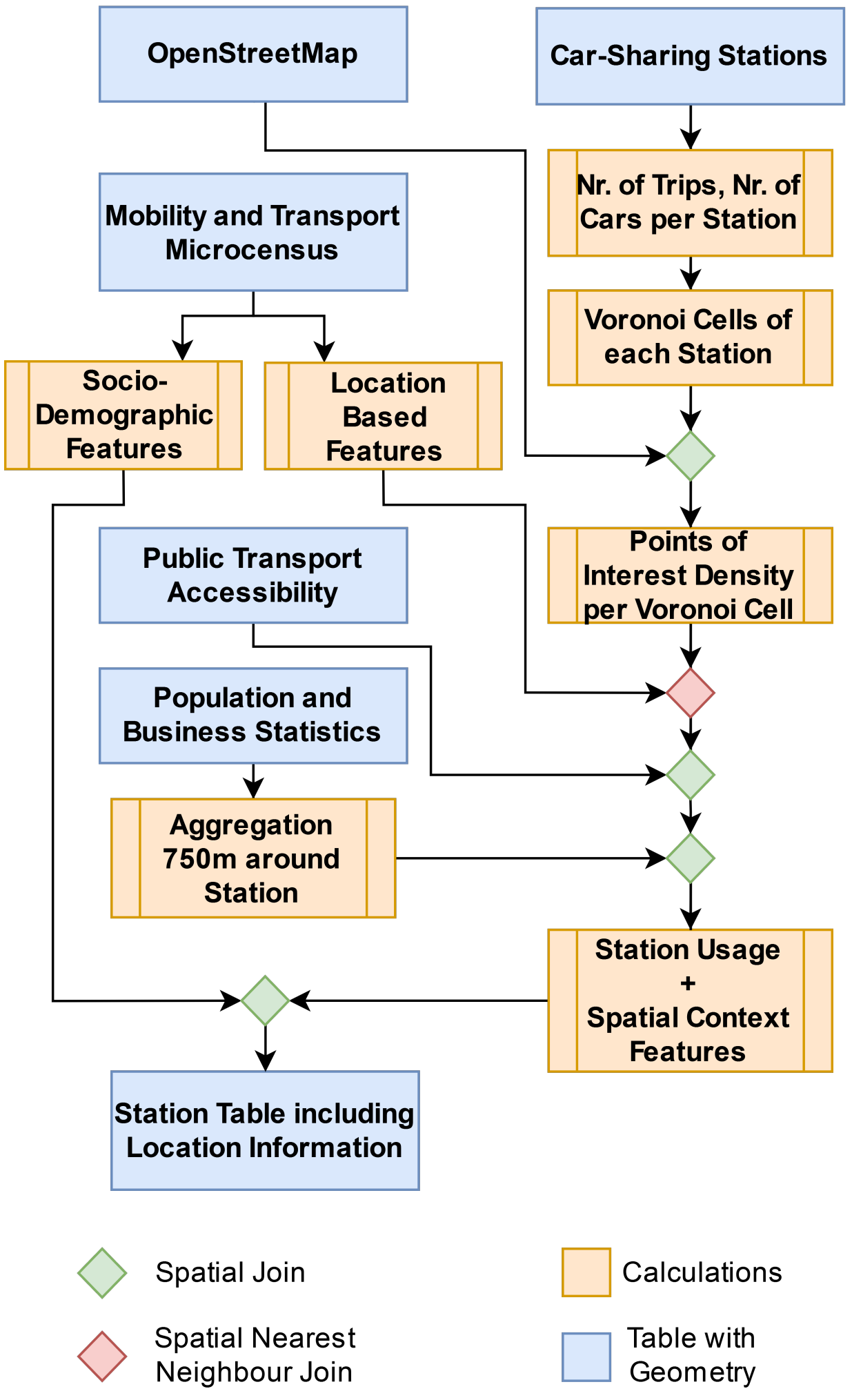

**Figure 1:** Flowchart of the data fusion procedure.

Using a spatial join, we added the public transport accessibility features at the location of the car-sharing station. Later, we added population and business census data by aggregating the values within 750m around each station. This process is explained visually in Figure 2c. The bandwidth of 750m was determined empirically by observing the predictive performance of the Random Forest model for varying bandwidths (250m, 500m, 750m, 1km, 2km, and 5km). Furthermore, we added features that measure the internal competition with other stations. Specifically, we calculated the number of competing stations and the number of competing cars within 1km of each station. Finally, we added aggregated personal features within a 1km distance around each station from the Microcensus dataset (income, number of cars per household). For 95 (6.6%) stations, less than ten Microcenus households were within the radius. For these, the calculation was repeated by increasing the distance until ten Microcensus households were found.

The final dataset consists of station-wise socio-demographic features, geographic features, and car-sharing-related features. Table 5 in Appendix B lists all available features with their types and basic statistics.

## 4. Methods

We compare multiple spatially-explicit models (including both linear models and non-linear models) for car-sharing demand prediction, with spatially-implicit models used as the baseline for comparison. These models are evaluated in terms of prediction accuracy, model complexity, ability to account for spatial heterogeneity and dependence, and interpretability. For interpretability, we focus on understanding the influence of features on demand prediction. In particular, we leverage SHAP values to interpret the RF model, and analyze GWR and MGWR coefficients to explain spatially varying predictor-response relationships. In the following, we describe the included models and the evaluation metrics for the model comparison.

### 4.1. Models

We compare five main models that are suitable for multivariate analysis: global Ordinary Least Squares regression, global Random Forest, Geographically Weighted Regression (GWR) [Brunsdon et al., 1998], Multi-Scale GWR [Fotheringham et al., 2017] and Geographical Random Forest [Georganos et al., 2021]. These models and our implementation details are described in the following.

#### *4.1.1. Ordinary Least Squares (OLS) regression*

The OLS regression model is defined by the following relationship between predictors and target variable [Trevor, 2009]:

$$y_i = \sum_j X_{ij}\beta_j + \varepsilon_i$$

where $X_{ij}$ is the jth predictor variable of the observation $i$, $\beta_j$ is the jth coefficient, $\varepsilon_i$ is the error term, and $y_i$ is the response variable.

#### *4.1.2. Geographically Weighted Regression (GWR)*

Compared to global models, the GWR model allows parameters to vary spatially. For each station, a local regression model is fitted. The model can be formulated as follows [Brunsdon et al., 1998]:

$$y_i = \sum_j \beta_j(u_i, v_i)X_{ij} + \varepsilon_i$$

The coefficient $\beta_j$ is spatially dependent on the location $(u_i, v_i)$ of the oberservation $i$. The concept of GWR follows Tobler's first law of geography [Tobler, 1970]: In the local regression model, spatially closer stations have a stronger influence on prediction. A bandwidth defines how many stations should be included in the local regression. The bandwidth can either be a *fixed* distance or an *adaptive* distance. In the latter case, the number of included neighbors is fixed and determines the distance radius. The weighting follows a kernel function (e.g., gaussian, bisquare, or exponential), depending on an estimated bandwidth. The optimal bandwidth is estimated using the golden-section search algorithm, either by minimizing the Corrected Akaike Information Criterion (AICc) score of the model or by cross-validation. The model calibration can be conducted using weighted least squares [Fotheringham et al., 2017]. We used the Python package *mgwr* for the implementation [Oshan et al., 2019].

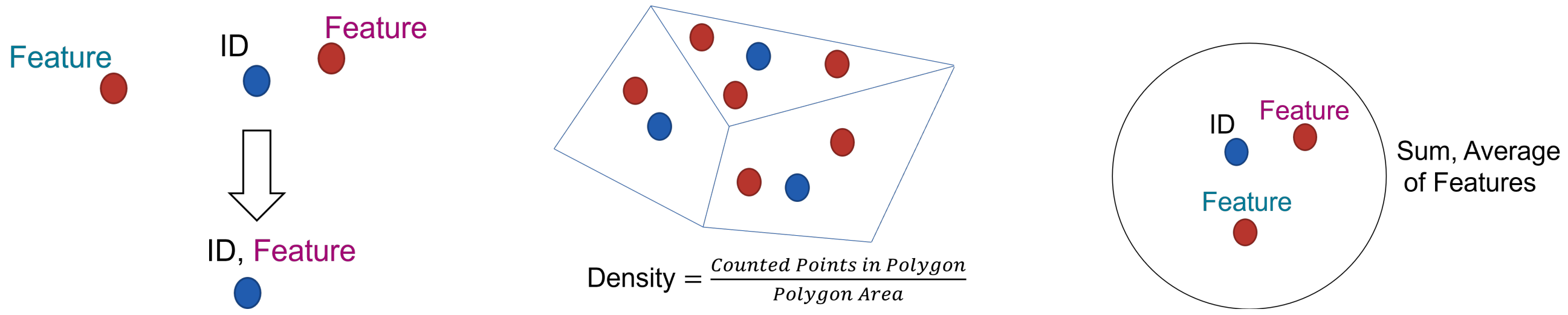

(a) Spatial nearest neighbour join (b) Aggregation using Voronoi polygons. (c) Aggregation within radius

**Figure 2:** Used spatial aggregation methods.

#### 4.1.3. Multi-Scale Geographically Weighted Regression (MGWR)

In GWR, all features are constrained to vary at the same spatial scale. Since not all features are necessary to exhibit local relationships with the response variable at the same scale, a global bandwidth might not be optimal for all features. MGWR is an extended version of GWR, where an optimal bandwidth is estimated for each feature. The model can be formulated as follows [Fotheringham et al., 2017]:

$$y_i = \sum_j \beta_{bwj}(u_i, v_i)X_{ij} + \varepsilon_i$$

Here, $\beta_{bwj}$ refers to the bandwidth used for calibrating the jth feature. We chose the bandwidths with the golden-section search algorithm, which minimizes the AICc criterion of the model. The MGWR model is calibrated using a back-fitting algorithm, which maximizes its expected log-likelihood. This procedure is described in detail in [Fotheringham et al., 2017]. Again, the *mgwr* Python package is used [Oshan et al., 2019]. Note that the current methodology developed for MGWR does not support *inference* on test data. Thus, we report only in-sample performance and fall back to GWR for out-of-sample experiments.

#### 4.1.4. Random Forest (RF)

The random forest regression algorithm combines a certain number of regression trees into an ensemble. Each tree is trained independently from the other, and the final prediction is made by taking the average prediction from each tree. Formally, the prediction of the random forest, $\hat{f}(x_i)$, is computed with the following formula [Breiman, 2001]:

$$\hat{f}(x_i) = \frac{1}{N}\sum_{n=1}^{N} f_b(x_i) \qquad y_i = \hat{f}(x_i) + \varepsilon_i$$

where $x_i$ is the feature vector, $N$ denotes the total number of independent trees in the forest, and $f_b(x_i)$ refers to the prediction of a single tree $b$. Random forest regression uses a bootstrapping scheme: Each tree is trained on a subset of the features and data (drawn with replacement). The benefit of bootstrapping is that outliers influence the prediction much less than using the total dataset for each tree. These two sources of randomness decrease overfitting and improve the robustness of the prediction [Breiman, 1996, 2001]. Each individual tree in a Random Forest is built by recursively partitioning the data into subsets and then selecting the best feature and split point at each node based on Gini impurity criteria. Random Forest regression achieved the highest accuracy in car-sharing prediction tasks in existing papers [Cocca et al., 2020].

Furthermore, we distinguish a special type of RF where the geographic coordinates are used as input features, treating the projected x and y coordinates as individual features. An illustrative example of this approach can be found in [Hengl et al., 2018]. This is motivated by the hypothesis that the random forest can capture spatial effects from the geographic coordinates.

#### 4.1.5. Geographical Random Forest (GRF)

The RF is a global concept that might not address spatial heterogeneity sufficiently besides the use of spatial input features. The GRF is an extension of RF that uses local sub-models. Similarly to GWR, a local model for each station is created based on the closest stations. The model can be formulated as follows [Georganos et al., 2021]:

$$y_i = \hat{f}(x_i, u_i, v_i) + \varepsilon_i$$

where $\hat{f}(x_i, u_i, v_i)$ refers to the prediction of the random forest spatially dependent on the location $(u_i, v_i)$ of the observation $i$. As in GWR, a local model is fitted for each station based on the nearest stations. At test time, the model fitted for the spatially *closest* sample is used for prediction. Note that this method is very computationally expensive since the number of models scales with the number of samples. We developed a fast implementation of the method in Python and published our code on GitHub[2]. In contrast to the original proposal of the method, we do not fuse the local model with a global RF since we want to provide a fair comparison of local and global methods without confounding the two.

## 4.2. Feature selection

To ensure reliable estimates of coefficients in the linear models (i.e., OLS regression, GWR, and MGWR), collinearity issues should be addressed [Wheeler and Tiefelsdorf, 2005]. Thus we used LASSO regression to select a subset of features. LASSO regression shrinks the regression coefficients by imposing a penalty on their size [Trevor, 2009]. The LASSO estimate is defined as:

$$\hat{\beta}^{lasso} = argmin_{\beta} \left\{ \frac{1}{2} \sum_{i=1}^{N} (y_i - \beta_0 - \sum_{j=1}^{p} x_{ij}\beta_j)^2 + \lambda \sum_{j=1}^{p} \left|\beta_j\right| \right\}$$

where $y_i$ is the dependent variable of the i-th sample, $x_{ij}$ are the independent variables, and $\beta_j$ are the estimated coefficients. The term $\lambda \geq 0$ refers to a complexity parameter that controls the amount of shrinkage. A larger value of $\lambda$ will lead to a greater shrinkage. Compared to other regularization techniques, the coefficients of unimportant features shrink to zero. We used the Python package *scikit-learn* for the implementation [Pedregosa et al., 2011]. Although LASSO reduces the number of features considerably, it picks correlated features arbitrarily during the feature selection. Thus, we also included manually selected features that have shown to be important for car-sharing demand prediction according to previous studies and empirical knowledge. In addition, local collinearity can occur even if no global issues exist. Therefore, the collinearity of the final set of predictors for the GWR model is checked using the local variance inflation factors (VIF) and condition number (CN) indices. Features that have strong collinearity are selectively removed based on their importance in predicting car-sharing demand. This procedure was repeated for each fold used for cross-validation and is also applied to the whole dataset used for inference (see Appendix Table 5).

## 4.3. Metrics for model comparison

We compared the model performances using a ten-fold cross-validation procedure. The following metrics were used:

### 4.3.1. Corrected Akaike Information Criterion (AICc)

The Akaike Information Criterion (AIC) is a commonly used model comparison and selection diagnostic that accounts for the trade-off between the model accuracy and complexity. However, AIC tends to favour overfitting models, especially if the sample size is small or when the number of fitted parameters is a moderate to a large fraction of the sample size [HURVICH and TSAI, 1989]. Since the fraction of our model parameters and the sample size is around 40, we use the corrected Akaike Information Criterion [Fotheringham et al., 2017]:

$$AICc = 2nln(\hat{\sigma}) + nln(2\pi) + n\frac{n + tr(S)}{n - 2 - tr(S)}$$

where $n$ is the local sample size, $\hat{\sigma}$ is the estimated standard deviation of the error term and $tr(S)$ is the trace of the hat matrix $S$.

### 4.3.2. Root Mean Squared Error (RMSE)

The RMSE refers to the root of the mean of squares from the residuals in a regression model. A smaller RMSE indicates a better fit for the model. The RMSE can be calculated as follows:

$$RMSE = \sqrt{\frac{\sum_{i=1}^{n}(y_i - \hat{y}_i)^2}{n}}$$

Thereby, $n$ is the total number of predicted values, $y_i$ is the true value of the target variable $i$, and $\hat{y}_i$ is the prediction of the target variable $i$.

### 4.3.3. Coefficient of determination ($R^2$)

The coefficient of determination represents the proportion of the variance that a regression model can explain using the independent variables. The coefficient refers to the goodness of fit. The following formula shows the calculation [Ludwig Fahrmeir, 2020]:

$$R^2 = \frac{\sum_{i=1}^{n}(\hat{y}_i - \bar{y})^2}{\sum_{i=1}^{n}(y_i - \bar{y})^2}$$

$\hat{y}_i$ refers to the predicted value of $i$, $y_i$ refers to the true value of $i$, $n$ to the number of predicted values, and $\bar{y}$ to the mean value of the prediction variable. A value of 0 means that the model can not explain the variance around the mean, whereas a value of 1 can completely explain the variance of the data.

### 4.3.4. Adjusted $R^2$

The adjusted $R^2$ is a refinement of the $R^2$ statistic. In addition to measuring the goodness of fit, it also considers the complexity of the model, thus providing a penalty for increased complexity. The metric is calculated using the formula below [Ludwig Fahrmeir, 2020]:

$$Adjusted\ R^2 = 1 - \frac{n-1}{n-p}(1 - R^2)$$

$n$ refers to the number of observations used for fitting the model and $p$ to the number of parameters in the model.

### 4.3.5. K-Fold Cross-Validation

K-Fold Cross-Validation is a technique for assessing regression models. It operates by partitioning the data into K subsets. The model undergoes training and evaluation K times through K iterations, with each subset acting as the validation set once. This process generates an average performance measure across iterations, estimating the model's generalization capability to new data. The mathematical representation of K-Fold Cross-Validation is defined as follows:

$$CV_K = \frac{1}{K}\sum_{i=1}^{K} L_i$$

[2] https://github.com/mie-lab/spatial_rf_python

$CV_K$ symbolizes the cross-validation score using K folds, and $L_i$ represents the performance metric computed on fold $i$, where the model is trained on all folds except $i$.

#### *4.3.6. Leave-One-Out cross-validation $R^2$ (LOOCV $R^2$)*

LOOCV is a specialized case of K-Fold Cross-Validation, where the number of folds K corresponds to the total available observations. Despite the computational costs associated with fitting and evaluating $n$ models, an efficient analytical solution exists for linear models when equipped with the hat matrix H. The LOOCV's mean-squared error (MSE) is expressed as [Trevor, 2009]:

$$MSE_{LOOCV} = \frac{1}{n}\sum_{i=1}^{n}\left[\frac{y_i - \hat{y}_i}{1 - H_{ii}}\right]^2$$

$MSE_{LOOCV}$ represents the out-of-sample mean-squared error, while $n$ represents the sample size, $y_i$ stands for the actual value and $\hat{y}_i$ represents the model's fitted value for the ith observation. The term $H_{ii}$ corresponds to the ith diagonal element of the hat matrix $H$ derived from the regression model. Then, the out-of-sample $R^2$ value can be computed using the formula below:

$$LOOCV\ R^2 = 1 - \frac{MSE_{LOOCV}}{Var(y))}$$

where $Var(y)$ represents the variance of the dependent variable.

#### *4.3.7. Residual Moran's I p-Value*

Moran's I is a spatial autocorrelation statistic used to assess the degree of spatial clustering of the regression residuals. It quantifies whether similar residuals are clustered or dispersed across a geographical area. If residuals have significant spatial autocorrelation, the regression model creates biased predictions in some geographical regions. The formula for Moran's I is [Goodchild, 1986]:

$$I = \frac{n}{W}\frac{\sum_{i=1}^{n}\sum_{j=1}^{n} w_{ij}(r_i - \bar{r})(r_j - \bar{r})}{\sum_{i=1}^{n}(r_i - \bar{r})}$$

Where $n$ refers to the number of observations, $r_i$ and $r_j$ are the regression models residuals of the observations $i$ and $j$, $\bar{r}$ is the mean of all residuals, $w_{ij}$ is the spatial weight between the units $i$ and $j$, and $W$ is the sum of all spatial weights. Monte Carlo simulation is employed to determine the significance of Moran's I index by generating numerous random spatial permutations of the data while maintaining the spatial structure. By comparing the observed Moran's I value with the distribution of values obtained from the simulations, the p-value can be calculated. This p-value indicates the likelihood of observing a similar Moran's I value purely by chance. If the p-value is smaller than 0.05, it suggests that the observed spatial pattern is statistically significant and not due to random chance.

### 4.4. Model interpretation methods and statistical inference

In addition to evaluating our models' goodness of fit and predictive accuracy, our objective is to draw insights about the driving factors of car sharing demand from them. This section introduces a method we employed to analyze the impact of individual features on predictions for the RF model. Furthermore, we describe the methodology we used to determine the significance of features in the Geographically Weighted Regression models.

#### *4.4.1. SHapley Additive exPlanations*

The SHapley Additive exPlanations (SHAP) method is a powerful framework for explaining predictions made by machine learning models. Based on cooperative game theory, SHAP offers a systematic approach to measure the contribution of a specific feature to an individual prediction. In the context of this paper, a prediction refers to the car-sharing demand at a particular station. To quantify this contribution, SHAP uses a metric known as the SHAP value, which is computed for each observation and feature. By calculating the mean absolute SHAP values for each feature, we can determine the average importance of that feature in making predictions. Moreover, SHAP values identify whether a feature has an increasing or decreasing effect on a specific prediction. In our analysis, we use SHAP values to understand the influence of each feature on the Random Forest models' predictions. More details about the method can be found in Lundberg and Lee [2017].

#### *4.4.2. Estimating feature significance for linear geographically weighted models*

For linear geographically weighted models such as GWR and MGWR, which fit numerous local models, statistical inference is conducted individually for each submodel. A common approach is to use a classic t-test to assess the significance of each feature locally. However, this method introduces a challenge since multiple tests on the same hypothesis can result in a set of false positives, even if the observed relationship is only a product of random chance. To address this issue, we have used a method for calculating adjusted alpha values as described in da Silva and Fotheringham [2016], distinct from the standard alpha value of 0.05. These adjusted alpha values are then used as the threshold for rejecting the null hypothesis, which refers to a non-statistically significant relationship between the feature and the dependent variable.

# 5. Results

## 5.1. Method comparison

### 5.1.1. *In-sample model evaluation*

Table 1 presents in-sample adjusted R-squared and AICc values for various linear regression models fitted to the entire dataset. These metrics are widely recognized for evaluating model goodness-of-fit while accounting for model complexity. Additionally, we assessed spatial autocorrelation by calculating Moran's I p-values for the models' residuals.

The *OLS Regression* consistently underperforms other linear models in terms of both metrics. The significant Moran's I p-value suggests spatial autocorrelation in the OLS residuals, resulting in doubt on the reliability of its estimated coefficients. This spatial dependence likely arises from spatial heterogeneity, which *OLS Regression* fails to capture adequately. The results from *GWR* demonstrate the superiority of spatially-aware regression methods for modelling spatial heterogeneity. *GWR* exhibits superior adjusted R-squared and AICc values when compared to *OLS Regression*. Moreover, *GWR* effectively addresses spatial autocorrelation in the residuals, as the Moran's I p-value indicates. By individually modeling the spatial scale of each feature, *MGWR* achieves even better adjusted R-squared and AICc values, by capturing spatial heterogeneity with the highest precision.

To evaluate the potential of Random Forest in modeling spatial heterogeneity, we calculated Moran's I p-values for the residuals of each Random Forest model. The results indicate that the *Global Random Forest* struggles to model spatial heterogeneity, likely due to the absence of georeferenced features. Interestingly, including geocoordinates as features improves the model's ability to capture spatial heterogeneity, effectively removing the significant spatial dependence of the residuals as demonstrated in the *Global Random Forest with coordinates* model. Additionally, our implementation of the *Geographical Random Forest* demonstrates its capability in modeling spatial heterogeneity as well, as indicated by the highly insignificant Moran's I p-value.

### 5.1.2. *Out-of-sample model validation*

We report the out-of-sample results of the best models respectively for each method in Table 1. The metrics marked with "Out-of-Sample" were obtained in 10-fold cross-validation, where the held-out test data was neither used for feature selection nor for parameter tuning.

The best results in terms of out-of-sample RMSE and R-squared are obtained with a *Global Random Forest*, achieving a higher R-squared score and a lower RMSE than the linear models (see Table 1). As expected, including the (projected) geographic coordinates as covariates in the *Global RF with coordinates* model improves the results. This indicates that spatial heterogeneity exists, which can be better modeled by including geographic coordinates in a Random Forest model. In contrast, our implementation of the *Geographical Random Forest* is significantly inferior, with an R-squared score of 0.82, even outperformed by *OLS Regression*, which yields surprisingly good results. When tuning the bandwidth of the Geographical RF, we observed that the performance increased as we increased the bandwidth, indicating that the benefit of modeling locality is limited.

The out-of-sample performance of the best *GWR* model is almost on par with the *Global RF*. This model was tuned with the AICc criterion on the training data of each fold, with an exponential kernel and a distance-based fixed bandwidth of 39.6km on average over the ten folds. For *MGWR*, the theory and methodology for out-of-sample are still not matureFotheringham et al. [2017], Comber et al. [2022], and currently no openly available implementation is available. Therefore, our implementation does not support out-of-sample prediction. However, the LOOCV R-squared indicates the superior generalization ability of the *MGWR* model compared to *OLS Regression* and *GWR*. A direct out-of-sample prediction performance comparison between the Random Forest Models and *MGWR* is not possible since no closed-form solution of LOOCV for Random Forest exists, and other estimation methods are computationally infeasible.

Overall, the predictive performance is very high, considering the difficulty of the task. To give an intuition about the task difficulty and prediction accuracy, we visualize the distribution and correlation of ground truth demand and predicted demand in Figure 3. In this case, the predictions on the test data set by the *Global Random Forest* model are used. There are considerable outliers, i.e. stations with up to 500 car reservations per month, as shown in Figure 3a (they are omitted in Figure 3b for a better visualization). These outliers also affect the RMSE reported in Table 1. Figure 3b demonstrates that the ground truth and predicted car-sharing demand generally align well.

## 5.2. Model interpretability and spatial heterogeneity

For the sake of interpretability, we will first analyze the *Random Forest with coordinates* feature importance with SHAP values. Secondly, we will interpret and compare the spatially varying coefficient estimates from *GWR* and *MGWR*. Furthermore, we will plot the spatial distribution of selected coefficients for the *MGWR* model and interpret the results in relation to possible spatial influences. We will also analyze and discuss the spatial scale of features when fitting the MGWR model. Finally, we will identify non-linear relationships by comparing feature importance between non-linear and linear models. In all cases, the entire dataset will be used to fit the model, as opposed to cross-validation, in order to provide a comprehensive overview.

### 5.2.1. *Interpretability through SHAP values*

We utilize SHAP values to analyze the *Random Forest with coordinates* model's feature importance. The SHAP values are calculated for each prediction individually for a specific feature. Figure 4a shows the mean absolute SHAP values of the most important features over all predictions. The *Supply of Cars at a Station* is by far the most influential feature, followed by the *Public Transport Accessibility* and

| Algorithm | Fixed/Adaptive | Kernel | Adjusted $R^2$ | AICc | Out-of-Sample RMSE | Out-of-Sample $R^2$ | LOOCV $R^2$ | Residual Moran's I P-Value |
|---|---|---|---|---|---|---|---|---|
| OLS Regression | | | 0.878 | 1090.35 | 26.2092 | 0.8599 | 0.8730 | 0.032 |
| GWR | Fixed | Exponential | 0.897 | 978.98 | 25.3400 | 0.8682 | 0.8785 | 0.214 |
| MGWR | Adaptive | Bisquare | **0.908** | **776.36** | | | **0.8835** | 0.271 |
| Geographical Random Forest | Adaptive | Boxcar | | | 30.6045 | 0.8225 | | 0.458 |
| Global Random Forest | | | | | 25.7265 | 0.8705 | | 0.039 |
| Global RF with coordinates | | | | | **24.7242** | **0.8767** | | 0.121 |

**Table 1**
Model performance for station-based car-sharing demand prediction. Best performances are marked in bold. The global random forest model with geographic coordinates achieves the best out-of-sample R-squared score and the lowest RMSE, whereas MGWR achieves the lowest AICc score.

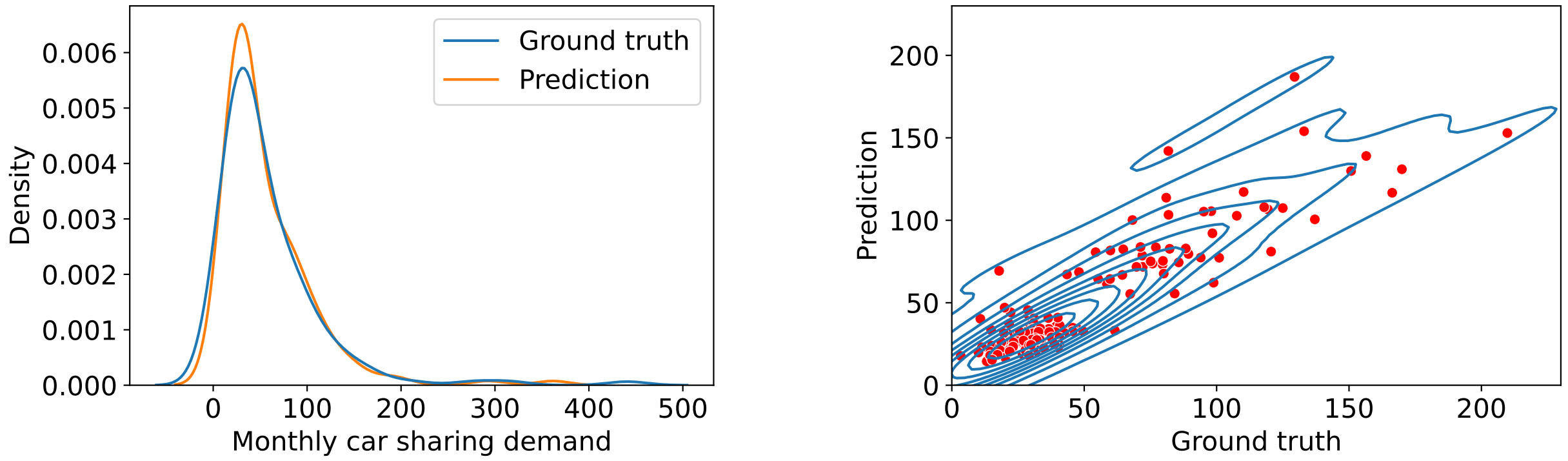

(a) Distribution of monthly car-sharing demand

(b) True vs. predicted demand

**Figure 3:** Comparing the predicted to the true monthly car-sharing demand. One sample is one station. (a) The distribution of the station-wise predictions aligns with the ground truth distribution. (b) The correlation between ground truth and prediction is high, and high-demand outliers with more than 150 trips per month are predicted accordingly.

the *Density of Public Points of Interest*. Figure 4b provides insights into how those features influenced individual predictions: High values of a feature are represented by red dots, while lower ones are marked in blue. We can read the SHAP value for each dot's prediction from the x-axis. The most important features are related to the centrality of a car-sharing station, where stations in densely populated regions generate more trips. Furthermore, the Random Forest uses the projected coordinates (especially the Y-coordinate) of the station's location as an important predictor. High values of Y-Coordinate (representing stations located in the north of Switzerland) yield higher prediction of the number of trips, indicated by the positive SHAP values.

#### 5.2.2. *Analysis of GWR and MGWR coefficients*

Table 2 and Table 3 refer to coefficient estimates of the most significant features from the *GWR* and *MGWR* models (A complete list of feature coefficient estimates is shown in Appendix Table 6 and Table 7). The coefficient estimates vary over space for both methods since a local geographically weighted regression model is fitted for every station. While the mean and median of the coefficients over all stations are related to the feature's importance, the standard deviation provides information about how the coefficients vary spatially. Further, the percentage of significant estimations over all local models is reported for each feature. Coefficient estimations from *MGWR* are generally more accurate by modelling each feature's spatial scale individually. However, *GWR* estimated surprisingly similar coefficient values to *MGWR*, making it still a suitable candidate for modelling spatially-varying relationships, especially in cases where out-of-sample prediction is also needed.

#### 5.2.3. *Analyzing spatial heterogeneity through MGWR coefficient mapping*

To gain a more complete picture of spatial variations of the *MGWR* parameters, we visualize the spatial distribution of selected coefficients in Figure 5. They were estimated using a bisquare kernel function with an adaptive bandwidth. Each coloured polygon on the map corresponds to the Voronoi area of a single station. First, Figure 5a shows the estimated *MGWR* coefficients for the feature *Population* (864 neighbours), which refers to the number of inhabitants within 750m around a car-sharing station. A higher population results in more car-sharing demand, eventually through a higher number of potential customers. The coefficients in the northeast of Switzerland (white areas) are the lowest: The region includes Zurich, which is the largest city in Switzerland and has, by far, the highest number of car-sharing stations. The homogenous population distribution in Zurich likely drives the lower coefficient estimation for local models in the northeast. That is because each local model contains predominantly stations in the city with a similar population. On the other hand, the northwest (dark red) has the highest coefficients, which can be explained by the heterogeneous population structure. Each local model contains stations from dense cities and rural regions so that the feature can explain a larger proportion of the target variable in that area.

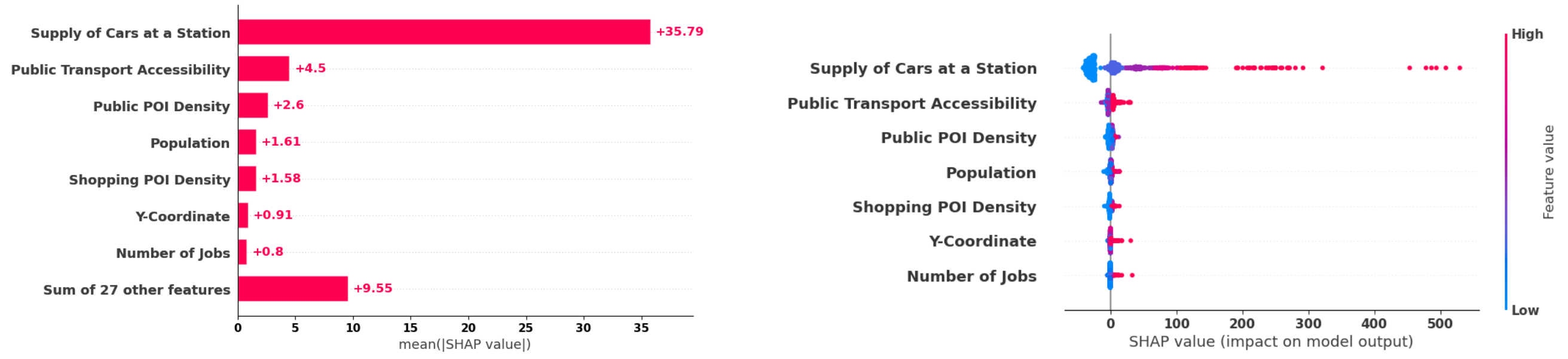

(a) Mean Absolute SHAP Values

(b) Individual SHAP Values

**Figure 4:** Feature importance in the *Global Random Forest with coordinates* model explained by SHAP values.

| Feature | Mean | $\sigma$ | Min | Median | Max | Mean t-value $\pm\sigma$ | Significant Estimates [%] |
|---|---|---|---|---|---|---|---|
| Supply of Cars at a Station | 0.906 | 0.043 | 0.655 | 0.894 | 0.967 | 62.77 $\pm$ 7.50 | 100.00 |
| Population | 0.113 | 0.027 | 0.062 | 0.107 | 0.187 | 4.41 $\pm$ 1.18 | 94.86 |
| Accommodation POI Density | -0.052 | 0.01 | -0.085 | -0.049 | -0.042 | -2.98 $\pm$ 0.72 | 80.19 |
| Public Transport Accessibility | 0.043 | 0.014 | 0.004 | 0.041 | 0.085 | 2.72 $\pm$ 0.86 | 59.62 |
| Number of Competing Cars | -0.054 | 0.017 | -0.082 | -0.055 | -0.013 | -2.44 $\pm$ 0.83 | 57.19 |
| Shopping POI Density | 0.036 | 0.019 | -0.021 | 0.041 | 0.07 | 2.19 $\pm$ 1.0 | 41.56 |
| Distance to Large Supermarket | -0.034 | 0.019 | -0.073 | -0.032 | 0.005 | -1.88 $\pm$ 1.15 | 29.33 |

**Table 2**
Most significant local coefficient estimates obtained from GWR model with a fixed bandwidth of 40.8 km and exponential kernel. The bandwidth was optimized using AICc.

Secondly, car-sharing stations with better *Public Transport Accessibility* tend to have more trips, as shown in Figure 5b. The feature's large bandwidth and the minimal spatial variance in the coefficients suggest that it impacts car-sharing demand at a global scale. This phenomenon is likely attributed to the high connectivity and coverage of the Swiss public transportation system.

Besides analyzing the demographics, *MGWR* can be used for diagnosing car-sharing services. Figure 5c shows the most important feature, namely the *Supply of Cars per Station* operating at a local scale bandwidth. Car-sharing service providers try to optimize vehicles' placement across the stations, depending on demand and costs. The parameter estimations for the canton Tessin (white regions) are significantly lower than for the rest of Switzerland. It might be that the vehicle placement there is not optimized on the same level as for the rest of Switzerland.

Figure 5d presents the coefficient patterns of the feature *Distance to Large Supermarkets*. The map shows overall negative coefficients with low spatial variability, with slightly less negative coefficients in the east and south of Switzerland (light blue regions). One potential explanation for the negativity of the coefficients is that regions with higher distances to large supermarkets have a lower centrality, often with less robust public transport systems and higher car ownership rates, leading to lower demand for car-sharing services. Notably, the regions with less negative coefficient values align remarkably well with the Swiss mountain regions. This pattern might be caused by the Euclidean distance used in calculating the distance to the nearest large supermarket, which might differ greatly from the actual travel distance because of natural barriers like canyons and mountains. Therefore, the feature may contain more noise in mountainous regions, leading to smaller absolute coefficient values as it may not accurately reflect the true accessibility.

| Feature | Mean | $\sigma$ | Min | Median | Max | BW* (Neighbours) | BW* Median | Mean t-value $\pm\sigma$ | Significant Estimates [%] |
|---|---|---|---|---|---|---|---|---|---|
| Supply of Cars at a Station | 0.908 | 0.127 | 0.552 | 0.903 | 1.173 | 232 | 35.2 km | 30.75 $\pm$ 9.55 | 100.00 |
| Public Transport Accessibility | 0.047 | 0.001 | 0.045 | 0.047 | 0.049 | 1438 | 227.9 km | 4.12 $\pm$ 0.12 | 100.00 |
| Distance to Large Supermarket | -0.037 | 0.0007 | -0.039 | -0.038 | -0.035 | 1438 | 227.9 km | -3.01 $\pm$ 0.2 | 100.00 |
| Population | 0.096 | 0.028 | 0.067 | 0.089 | 0.164 | 864 | 87.2 km | 4.07 $\pm$ 1.57 | 94.37 |
| Shopping POI Density | 0.028 | 0.003 | 0.024 | 0.027 | 0.033 | 1438 | 227.9 km | 2.27 $\pm$ 0.24 | 65.18 |
| Accommodation POI Density | -0.029 | 0.004 | -0.036 | -0.027 | -0.026 | 1409 | 220.8 km | -2.14 $\pm$ 0.32 | 34.89 |
| Income | 0.025 | 0.023 | 0.0002 | 0.009 | 0.060 | 750 | 76.2 km | 1.38 $\pm$ 1.21 | 23.63 |

**Table 3**
Most significant local coefficient estimates were obtained from the MGWR model with an adaptive bandwidth and bisquare kernel. The bandwidth was optimized using LOOCV.

While *MGWR* provides the most accurate and fine-scale coefficient estimations, *GWR* is also capable of modeling spatial heterogeneity and reveals similar patterns, as shown in Appendix A (Figure 7).

#### 5.2.4. *Spatial scale of features*

*GWR* assumes that all predictors and the *Intercept* have broadly similar bandwidths, which can fail to capture the multi-scale spatial processes. For that reason, previous research outlined the superiority of *MGWR* over single bandwidth methods for many data sets [Comber et al., 2022]. We fitted all stations using an *MGWR* model to analyze the spatial scale of each feature's impact on the dependent variable (see Table 7 in the Appendix for the distribution of features' bandwidths). Most features have a global or near global adaptive bandwidth of over 1000 neighbours (the maximum number of neighbours is 1439). Three out of 33 features have local spatial scales with bandwidths below 600 neighbours. These three features are *Number of Workplaces Sector 1 (Agriculture, Mining)* (bandwidth: 108), *Supply of Cars at a Station* (bandwidth: 232), and *Distance to Postal Office* (bandwidth: 498). In addition, the *Intercept* (bandwidth: 623), the feature *Income* (bandwidth: 750), and the feature *Population* (bandwidth: 864) show a medium size or regional spatial scale.

The feature values for most predictors were calculated using several spatial aggregation methods, described in subsection 3.2. This could lead to complex spatial interactions since both model and aggregation methods are calculated using spatial bandwidths. For most spatial aggregated features, the model's bandwidth is large (Appendix Table 7) relative to the aggregation bandwidth (subsection 3.2). For them, the effect of the spatial scale of aggregation is negligible. An exception is the feature *Number of Workplaces Sector 1 (Agriculture, Mining)* aggregated using a 750m buffer bandwidth around each station. The model's spatial scale for that feature is 2.5 km in the densest urban region. Since this feature's exact interaction of both bandwidths is complex, the result should be interpreted with caution. Despite the lack of an exact interpretation of the spatial scale, we can be sure that the feature presents local patterns.

#### 5.2.5. *Comparison of SHAP values, GWR, and MGWR coefficients*

To understand the differences in feature contributions to the prediction between global and local methods, we compare the feature importance for the *Global Random Forest with coordinates* estimated using the SHAP values (Figure 4) to the *GWR* (Table 2) and the *MGWR* coefficients (Table 3). For all models, the feature *Supply of Cars at a Station* is the most crucial feature. A strong linear relationship exists between the number of cars placed at a station and the number of trips. Further, the features *Public Transport Accessibility*, *Population*, *Shopping POI Density* seem to be essential predictors in all models. In addition, the features *Number of Jobs* and *Public POI Density* are important predictors in the Random Forest model. In contrast, their significance and coefficient values are low in both GWR and MGWR models. This observation indicates potential non-linearity between these features and the target/response variable.

### 5.3. Case study: estimating the optimal vehicle supply at a station

Our framework can serve as a decision-support tool to assist in the planning of new car-sharing stations. As a demonstration, we conduct a case study on estimating the demand for new potential station locations with our trained models. However, the supply of cars at a car-sharing station is an indispensable feature for calculating a station's demand. The differences in supply between the stations cannot be explained by socio-demographics alone. Thus, we estimate the demand while varying the vehicle-supply input feature, to show if and how the models can be used as a decision-making tool to determine not only the demand but also the optimal supply of vehicles at a station.

We selected three locations in the canton of Zurich for potential new car-sharing stations. The locations differ in their spatial context; one station is located in the city centre, the second is in an intermediate region, and the last is in the countryside. Then, the demand for each station was predicted by varying the vehicles placed at each station, as shown in Figure 6. The RF and *GWR* models were used since *MGWR* does not support out-of-sample prediction.

While *GWR* is a linear model such that the predicted demand constantly increases with the supply of cars, the *Random Forest* saturates for a higher number of vehicles. The demand to which the model converges highly depends on the socio-demographic features; the city location converges to a much higher demand than the rural or intermediate station. We analysed the station's neighbourhood to validate if the convergence thresholds are realistic. The average station size and demand within 3km around each station were calculated (circle with red outline). Additionally, we compared the predicted demands with the largest station within this neighbourhood in the plot (circle).

Most stations have only a small number of cars; even the city stations neighbourhood has only two cars on average. The model can simulate a realistic saturation pattern for the city location since the largest station in the neighbourhood shows a very similar demand and supply of vehicles. This is remarkable considering that the *Random Forest* is a global model and does not know about the closest station nearby. Therefore, the model helps plan the new stations' supply of cars. Whether to place all vehicles at a single station or to arrange several smaller stations depends on the availability of infrastructure and design strategies.

On the other hand, the model overestimates the demand for a higher supply of cars in the intermediate and rural regions. This issue can be explained by the fact that from the total 1439 fitted stations, only 31 and 173 stations are located in rural and intermediate regions, respectively. Therefore,

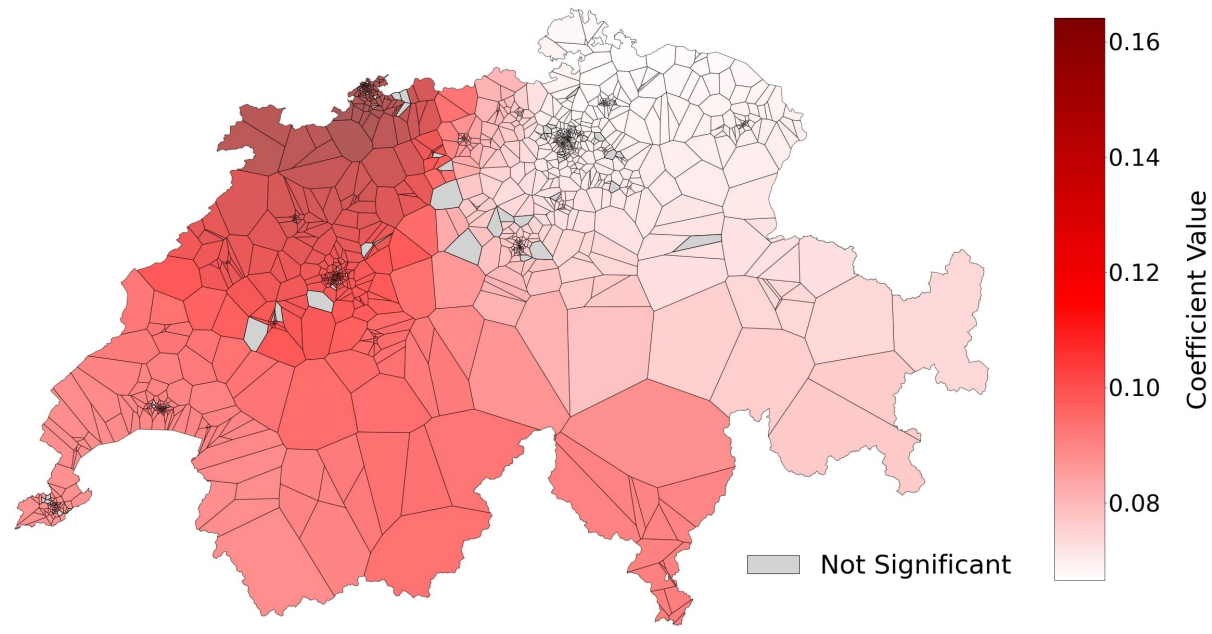

(a) Population (864 Neighbours)

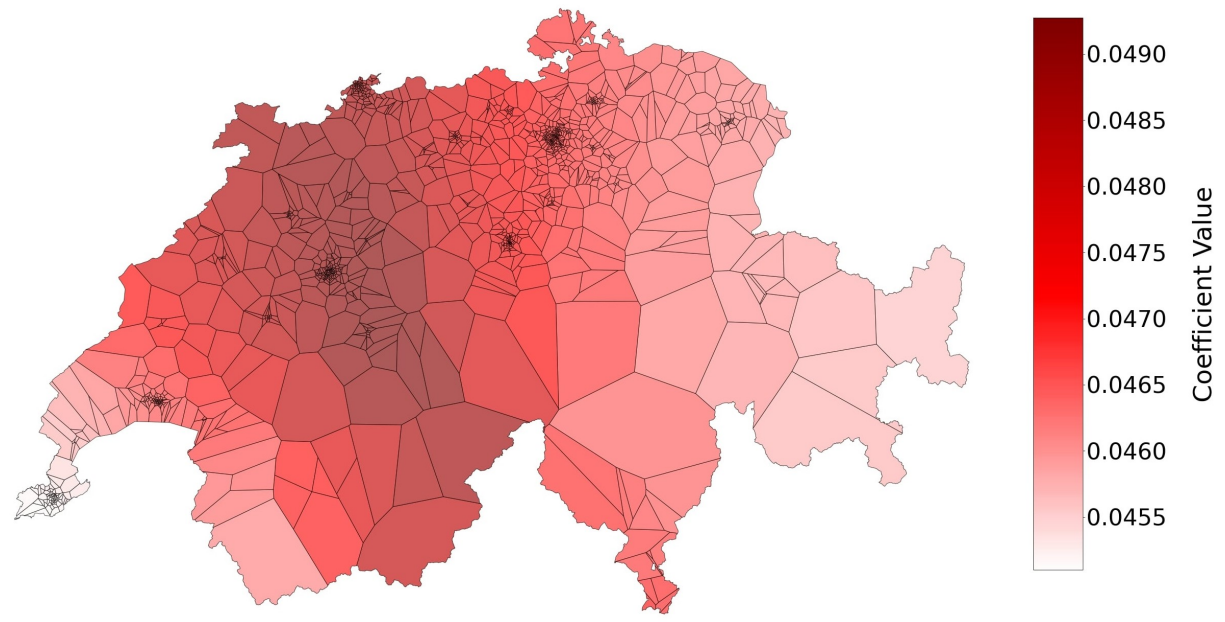

(b) Public Transport Accessibility (1438 Neighbours)

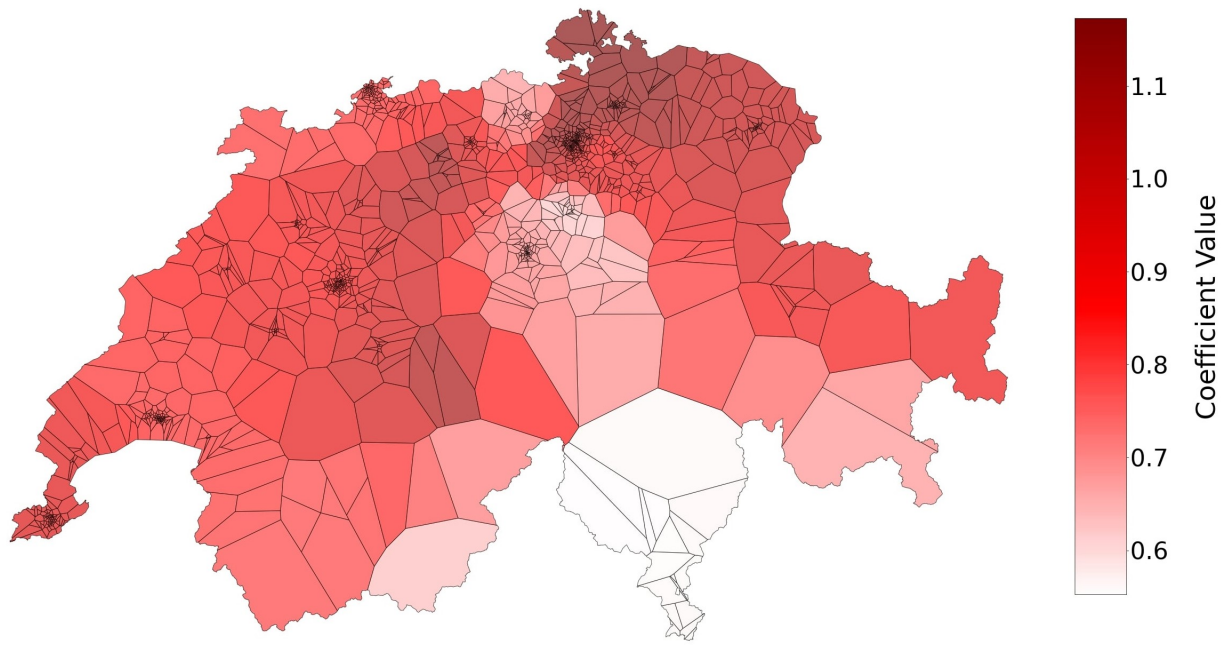

(c) Supply of Cars at a Station (232 Neighbours)

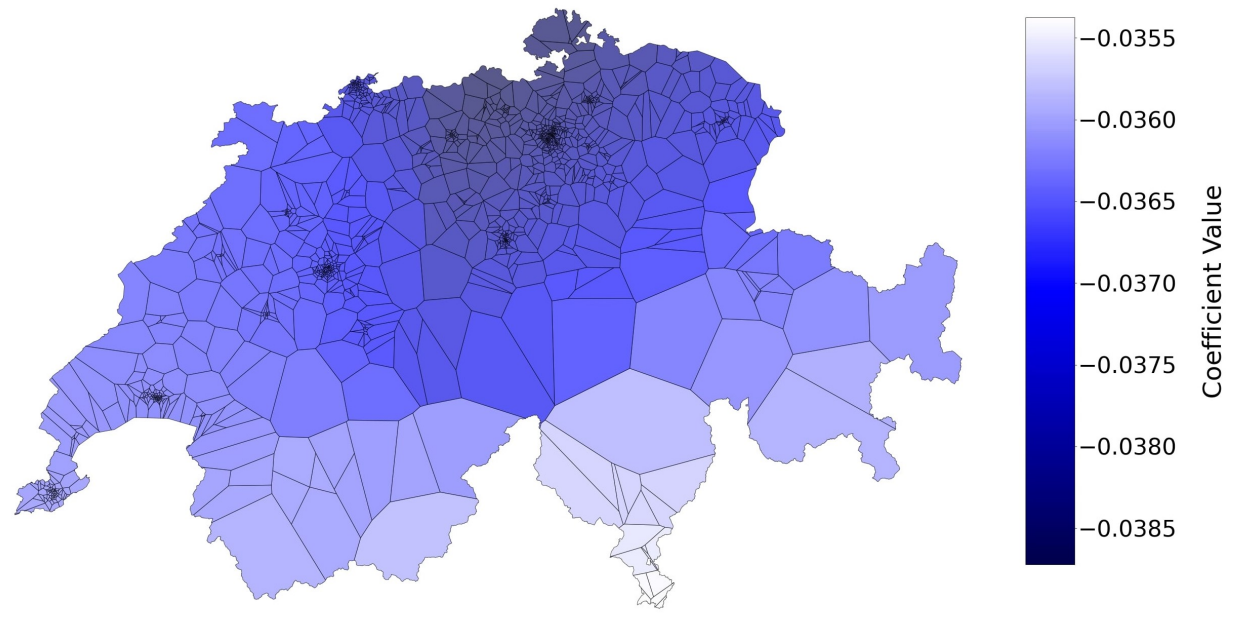

(d) Distance to Large Supermarket (1438 Neighbours)

**Figure 5:** Spatial variation of *MGWR* parameters (with the bandwidth in brackets).

the lack of training data can explain the model's overestimation in those regions.

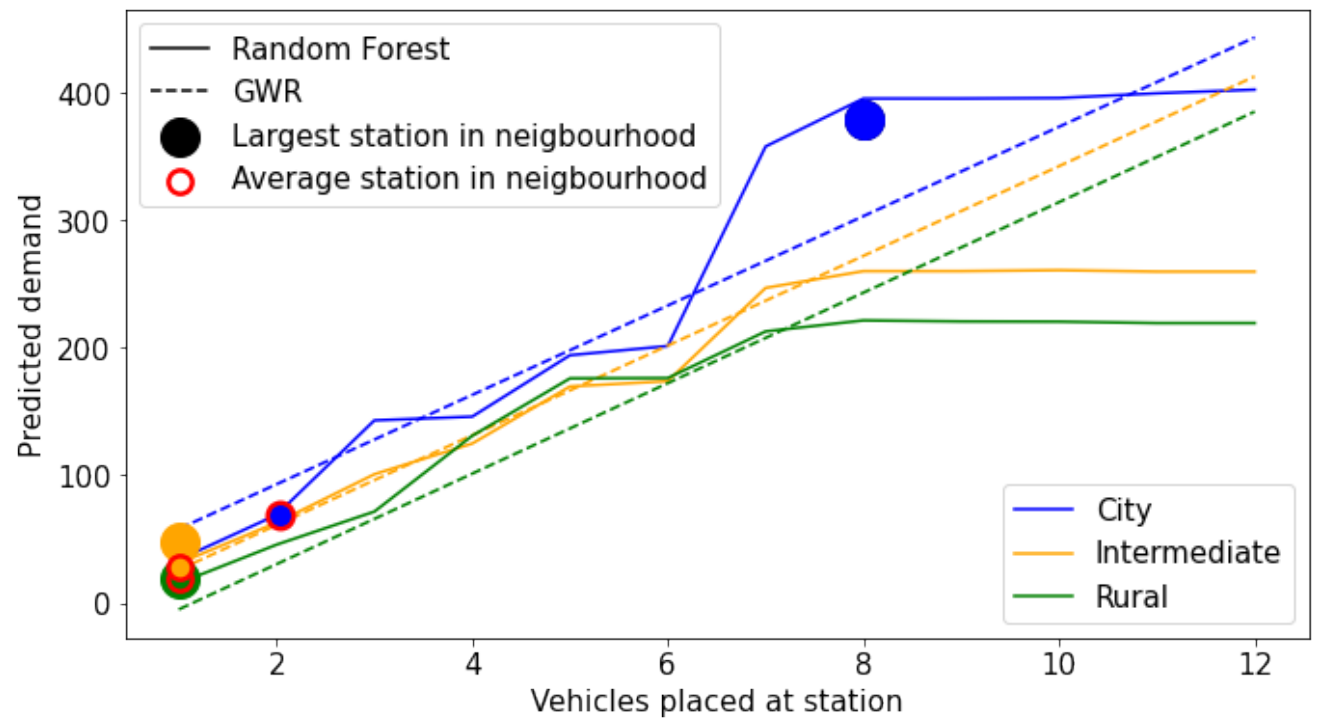

**Figure 6:** Predicted demand for *Random Forest* and *GWR* by varying the vehicle supply.

## 5.4. Ablation study: GWR bandwidth and kernel selection

To justify our parameter choice for the best GWR model in Table 1, we list the results for other variations of the model in Table 4. We systematically varied bandwidth types (adaptive or fixed), selection criteria, and kernel choices. In general, the parameters do not have a strong effect on the results, since all models achieve an R-squared score above 0.85. A fixed bandwidth performs better than an adaptive one, and the Exponential kernel is most suitable compared to a Gaussian or a Bisquare kernel. The best-performing *GWR* model has a fixed bandwidth of 39.6km on average and is fitted with an Exponential kernel using the AICc bandwidth selection criterion. This model achieved the best score among all GWR models evaluated in all three evaluation metrics, i.e., AICc, RMSE, and R-Squared.

# 6. Discussion

In summary, our study compared multiple spatially-explicit models and spatially-implicit models in predicting car-sharing demand in terms of their prediction accuracy and interpretability. In contrast to previous work [Zhu et al., 2019, Wang et al., 2021, Xu et al., 2018, Qiao et al., 2021] our study supports long-term decision-making instead of short-term operations and features spatial awareness.

Our work shows that a Random Forest model with geo-coordinates as input provides high predictive performance while accounting for the non-linear and spatial heterogeneity effects, rendering them particularly useful for planning new stations. In contrast, although MGWR does not offer out-of-sample prediction capability, it enables precise modelling of predictor-response relationships that vary over space and across different spatial scales. GWR strikes a balance between spatial interpretability and predictive ability, and, in contrast to MGWR, is well-established for out-of-sample prediction both in theory and implementation. The results of our comparative study offer useful guidelines for choosing suitable spatial explicit models for different applications.

| Algorithm | Fixed/Adaptive | Bandwidth Selection | Kernel | Adjusted $R^2$ | AICc | Out-of-Sample RMSE | Out-of-Sample $R^2$ | Residual Moran's I P-Value |
|---|---|---|---|---|---|---|---|---|
| GWR | Adaptive | AICc | Gaussian | 0.884 | 1067.23 | 26.4076 | 0.8588 | 0.432 |
| GWR | Adaptive | CV | Gaussian | 0.879 | 1083.10 | 26.1139 | 0.8606 | 0.116 |
| GWR | Adaptive | AICc | Bisquare | 0.892 | 1014.30 | 26.1953 | 0.8609 | 0.399 |
| GWR | Fixed | CV | Gaussian | 0.891 | 1005.23 | 26.0446 | 0.8616 | 0.434 |
| GWR | Adaptive | CV | Bisquare | 0.891 | 1016.60 | 25.7873 | 0.8636 | 0.437 |
| GWR | Adaptive | CV | Exponential | 0.887 | 1038.20 | 25.7822 | 0.8639 | 0.307 |
| GWR | Adaptive | AICc | Exponential | 0.893 | 1027.39 | 25.7459 | 0.8642 | 0.288 |
| GWR | Fixed | AICc | Gaussian | 0.891 | 1005.23 | 25.6667 | 0.8651 | 0.467 |
| GWR | Fixed | CV | Exponential | **0.898** | 981.19 | 25.4201 | 0.8676 | 0.173 |
| GWR | Fixed | AICc | Exponential | 0.897 | **978.98** | **25.3400** | **0.8682** | 0.214 |

**Table 4**
Analyzing the effect of GWR parameter settings on its performance. The overall scores are only affected to a small extent by the parameters. A fixed bandwidth and an exponential kernel yield superior performance.

In addition, we provided a detailed analysis of how spatial and socio-demographic features influence car-sharing demand prediction, leveraging interpretable methods that unveil non-linear and spatially heterogeneous effects. Our analysis demonstrated that spatial features such as POI density and public transport accessibility are valuable predictors of car-sharing demand. Furthermore, even socio-demographics such as population, income, and job availability play an important role in prediction, but with varying importance depending on the spatial scales.

While we consider a rich set of more than 72 features, the model may improve with further information. For example, the availability of car-sharing-user specific information was limited to age groups and gender in our dataset. In a real use case, the operator may conduct a survey on the mobility habits and other social factors in the area of several potential station locations, and the resulting information could additionally be used by the model. Our models are easy to implement with existing libraries in Python and R, and are applicable to data from other locations and other long-term demand applications, e.g., for bike sharing.

Furthermore, the specific dataset used in the study limits the generalisability of the results. For example, the user demand in a one-way car-sharing system may differ significantly, and our models may not be applicable since they were trained on return-trip data. Also, the situation in Switzerland is substantially different from other countries, given the long history of car-sharing and the strong importance of public transport. However, due to potentially strong local differences, the model should be retrained on existing stations of a car-sharing service in any case. In contrast to other studies on station demand prediction, we considered a particularly large dataset in terms of the number of stations and the spatial extent.

Finally, our analysis of SHAP values and MGWR / GWR coefficients showed the potential of these methods to identify influential factors. While global methods do not allow for detailed mapping of spatial variations of the predictor-response relationships such as in Figure 5, it is still possible to use these methods to model spatial heterogeneity by explicitly using spatial features as input, in our case, geographical coordinates. Further, we can investigate spatial patterns learned by these global models through the analysis of SHAP values. For example, we observe from the SHAP values in Figure 4b that the predicted demand increases for stations located in the north of Switzerland. However, MGWR and GWR are still superior in uncovering the heterogeneous spatial relationships between different predictors and the car-sharing demand. Overall, we highlight the importance of both local linear and global non-linear methods for their spatial interpretability and their ability to unveil more complex impact patterns respectively.

## 7. Conclusion

Players in the shared economy need to make logistic decisions every day, on the allocation and re-distribution of resources. The setup of new stations in a station-based system on the other hand involves larger investments and strongly affects the efficiency of short-term operations, as well as user satisfaction. In this work, we have shown how interpretable spatial regression methods can support the decision process with demand prediction and by identifying the effect of various influence factors.

In future work, it will be highly interesting to analyze how *changes* of the population features or spatial features affect the prediction of our model. For example, the COVID-19 pandemic has changed mobility behaviour substantially [Pan et al., 2020], including car sharing [Alonso-Almeida, 2022]. Our models could be amended with additional features to account for new influence factors arising from disruptive events, e.g., by including the prevalence of the home office as a feature. It is expected that our model trained before the pandemic overestimates the monthly demand during the pandemic, and that the spatial distribution of demand has changed. Considering that COVID-19 affected different regions to varying extents, it becomes even more important to use local methods that can account for spatial heterogeneity.

Finally, collaborating with car-sharing services could provide an actual validation of the model with real installations of new stations. In general, we hope that our work inspires new work to investigate long-term car-sharing behaviour which will support decision-making in such promising concepts for more sustainable future mobility.

## Acknowledgements

This research is part of the V2G4CarSharing project that is funded by the Swiss Federal Office of Energy (SFOE) through Grant SI/502344-01. We acknowledge the Mobility Cooperative for providing the car-sharing dataset in the context of this project. Furthermore, we would like to thank the Swiss Federal Statistics Office for providing the Mobility and Transport Microcensus, the Population and Household Statistics, and the Corporate Statistics data.

## CRediT author statement

**Dominik J. Mühlematter**: Conceptualization, Methodology, Software, Validation, Visualization, Writing – original draft. **Nina Wiedemann**: Conceptualization, Data curation, Methodology, Software, Visualization, Writing – original draft. **Yanan Xin**: Conceptualization, Funding acquisition, Methodology, Software, Project administration, Supervision, Writing – review & editing. **Martin Raubal**: Conceptualization, Funding acquisition, Project administration, Supervision, Writing – review & editing.

# Appendix

## A. Comparison of MGWR and GWR coefficient maps

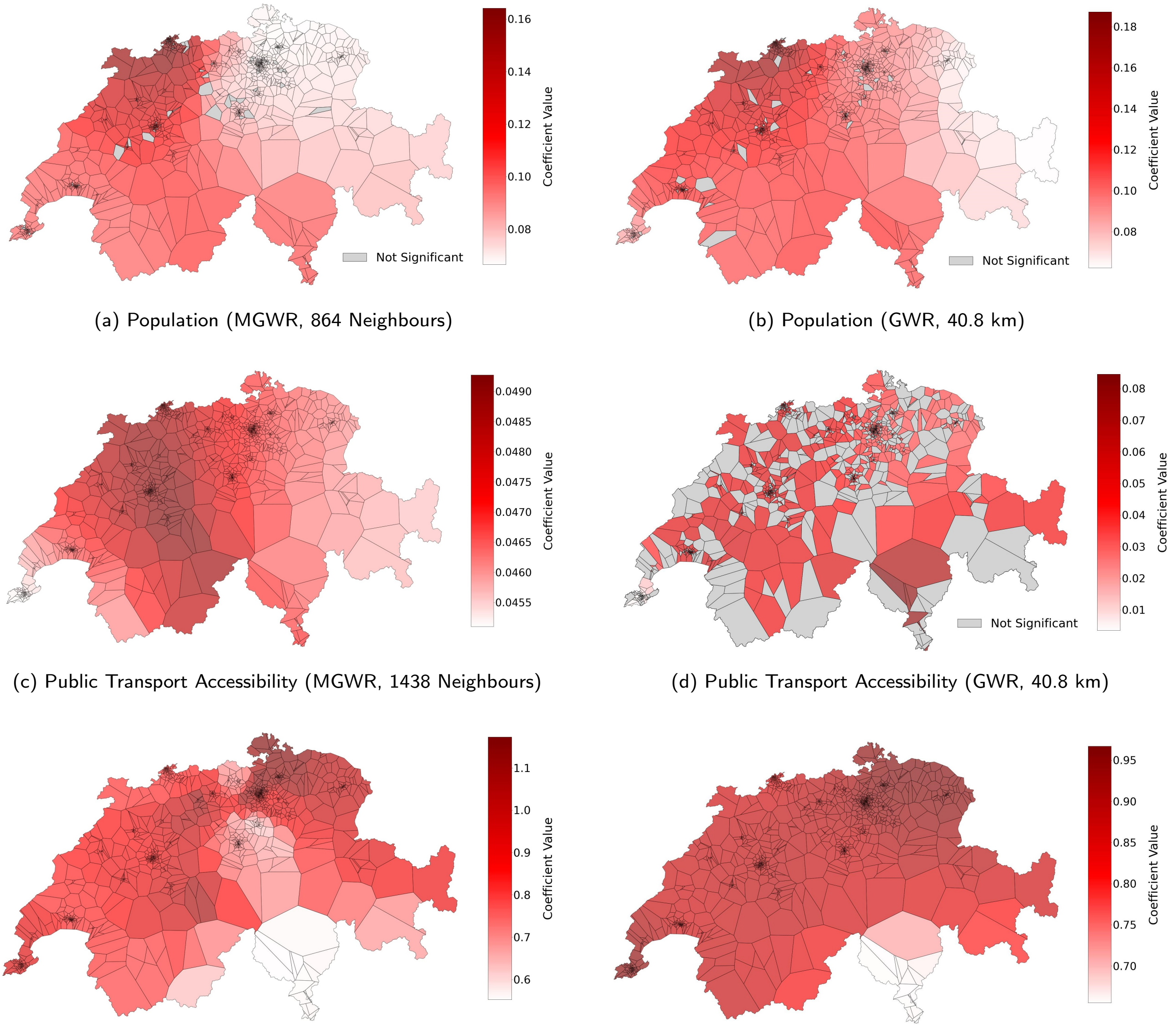

(a) Population (MGWR, 864 Neighbours)

(b) Population (GWR, 40.8 km)

(c) Public Transport Accessibility (MGWR, 1438 Neighbours)

(d) Public Transport Accessibility (GWR, 40.8 km)

(e) Supply of Cars at a Station (MGWR, 232 Neighbours)

(f) Supply of Cars at a Station (GWR, 40.8 km)

**Figure 7:** Comparison of GWR and MGWR parameter maps for some of the most significant features. GWR reveals very similar spatial patterns to MGWR. However, the significance of features can change due to differences in spatial scale (e.g., Public Transport Accessibility).

## B. All selected features

| Feature Name | Unit | Type | Mean | Standard Deviation | Selected By |
|---|---|---|---|---|---|
| Station's Voronoi Area | $km^2$ | Cardinal | 26.31 | 88.62 | LASSO |
| Public POI Density | $\#/km^2$ | Cardinal | 6.77 | 11.48 | LASSO |
| Health POI Density | $\#/km^2$ | Cardinal | 5.07 | 13.30 | Not selected |
| Leisure POI Density | $\#/km^2$ | Cardinal | 7.08 | 10.89 | Not selected |
| Food POI Density | $\#/km^2$ | Cardinal | 21.77 | 53.67 | Not selected |
| Accommodation POI Density | $\#/km^2$ | Cardinal | 3.3 | 7.07 | LASSO |
| Shopping POI Density | $\#/km^2$ | Cardinal | 36.42 | 89.54 | LASSO |
| Bank POI Density | $\#/km^2$ | Cardinal | 3.3 | 6.71 | Not selected |
| Tourism POI Density | $\#/km^2$ | Cardinal | 4.69 | 8.31 | Not selected |
| Average Distance to Car-Sharing Station | m | Cardinal | 865.51 | 1152.55 | Manually |
| Agglomeration Size Category 2000 | 0(lowest) - 6(highest) | Ordinal | 2.19 | 1.40 | Not selected |
| Spatial structure 2000 | 1(lowest) - 7(highest) | Ordinal | 2.04 | 1.37 | Not selected |
| Aggregated spatial structure 2000 | 0(lowest) - 4(highest) | Ordinal | 1.68 | 0.91 | Not selected |
| Travel Time to Agglomerations by Car | 1(lowest) - 4(highest) | Ordinal | 1.53 | 0.71 | Manually |
| Travel Time to Agglomerations by Public Transport | 1(lowest) - 4(highest) | Ordinal | 1.88 | 1.08 | LASSO |
| Distance to Restaurant | km | Cardinal | 0.43 | 0.33 | LASSO |
| Distance to Museum | km | Cardinal | 2.68 | 2.63 | LASSO |
| Distance to Theatre | km | Cardinal | 2.68 | 2.63 | Not selected |
| Distance to Hospital | km | Cardinal | 3.86 | 3.13 | LASSO |
| Distance to Postal Office | km | Cardinal | 0.88 | 0.40 | LASSO |
| Distance to Pharmacy | km | Cardinal | 1.45 | 1.93 | Not selected |
| Distance to Bank | km | Cardinal | 1.17 | 0.75 | Not selected |
| Distance to Cinema | km | Cardinal | 3.88 | 3.79 | Not selected |
| Distance to School | km | Cardinal | 1.28 | 1.75 | Not selected |
| Distance to Doctor | km | Cardinal | 0.82 | 0.97 | LASSO |
| Distance to Café/Pub | km | Cardinal | 1.52 | 1.78 | Not selected |
| Distance to Library | km | Cardinal | 5.07 | 6.23 | Not selected |
| Distance to Large Supermarket | km | Cardinal | 5.47 | 14.72 | LASSO |
| Distance to Small Supermarket | km | Cardinal | 1.35 | 1.35 | Not selected |
| Distance to Small Shop | km | Cardinal | 2.26 | 2.42 | LASSO |
| Distance to Large Shop | km | Cardinal | 1.22 | 1.05 | Not selected |
| Distance to Public Transport Station | km | Cardinal | 0.3 | 0.16 | LASSO |
| Public Transport Accessibility | 0(lowest) - 4(highest) | Ordinal | 3.21 | 0.89 | LASSO |
| Context of Living | 1(Urban) - 3(Rural) | Ordinal | 1.16 | 0.42 | Not selected |
| Number of Workplaces (All Sectors) | Jobs | Cardinal | 1118.68 | 1085.19 | Not selected |
| Number of Workplaces Sector 1 (Agriculture, Mining) | Jobs | Cardinal | 9.41 | 10.69 | LASSO |
| Number of Workplaces Sector 2 (Industry) | Jobs | Cardinal | 171.88 | 91.36 | Not selected |
| Number of Workplaces Sector 3 (Service) | Jobs | Cardinal | 1042.44 | 1043.87 | Not selected |
| Number of Jobs Sector (all sectors) | Jobs | Cardinal | 7399.47 | 8617.68 | LASSO |
| Number of Jobs Sector 1 (Agriculture, Mining) | Jobs | Cardinal | 12.45 | 19.58 | Not selected |
| Number of Jobs Sector 2 (Industry) | Jobs | Cardinal | 818.45 | 942.11 | Not selected |
| Number of Jobs Sector 3 (Service) | Jobs | Cardinal | 6634.01 | 8283.26 | Not selected |
| Population | Inhabitants | Cardinal | 9158.09 | 6437.30 | LASSO |
| Proportion of Inhabitants with Swiss Nationality | Swiss/Total Inhabitants | Cardinal | 0.7 | 0.11 | Not selected |
| Proportion of Swiss Born Inhabitants | Swiss Born Inhabitants/Total Inhabitants | Cardinal | 0.64 | 0.11 | LASSO |
| Proportion of Female Inhabitants | Female Inhabitants/Total Inhabitants | Cardinal | 0.51 | 0.02 | LASSO |
| Average Age | 6.6(lowest) - 10.6(highest) | Cardinal | 9.13 | 0.42 | LASSO |
| Number of Households | Households | Cardinal | 4479.39 | 3296.55 | LASSO (manually removed) |
| Average Household Size | Inhabitants | Cardinal | 2.43 | 0.40 | Manually |
| Supply of Cars at a Station | Cars | Cardinal | 1.88 | 1.97 | LASSO |
| Number of Competing Stations | Stations | Cardinal | 3.52 | 4.25 | LASSO (manually removed) |
| Number of Competing Cars | Cars | Cardinal | 7.9 | 10.83 | LASSO |
| Cars per Household | Cars | Cardinal | 1.01 | 0.35 | Manually |
| Parking per Household | Parking Places | Cardinal | 1.13 | 0.56 | Not selected |
| Motorbikes per Household | Motorbikes | Cardinal | 0.14 | 0.09 | LASSO |
| Bicycles per Household | Bicycles | Cardinal | 1.86 | 0.53 | Manually |
| Spatial Structure: Center | - | Binary | 0.54 | 0.50 | Not selected |
| Spatial Structure: Suburban Municipalities | - | Binary | 0.27 | 0.44 | Not selected |
| Spatial Structure: High-Income Municipalities | - | Binary | 0.04 | 0.20 | LASSO |
| Spatial Structure: Peri-Urban Municipalities | - | Binary | 0.06 | 0.24 | Not selected |
| Spatial Structure: Touristic Municipalities | - | Binary | 0.01 | 0.09 | Not selected |
| Spatial Structure: Industrial and Tertiary Municipalities | - | Binary | 0.04 | 0.19 | Not selected |
| Spatial Structure: Rural Commuter Town | - | Binary | 0.02 | 0.13 | LASSO |
| Spatial Structure: Agrarian-Mixed Municipalities | - | Binary | 0.03 | 0.16 | Not selected |
| Rural Municipality without Urban Characteristics | - | Binary | 0.06 | 0.24 | Not selected |
| Agglomeration Core Municipality (Core City) | - | Binary | 0.51 | 0.50 | Not selected |
| Agglomeration Core Municipality (Primary Core) | - | Binary | 0.22 | 0.42 | Not selected |
| Agglomeration Core Municipality (Secondary Core) | - | Binary | 0.07 | 0.25 | Not selected |
| Municipality of Agglomeration Zone | - | Binary | 0.08 | 0.27 | Not selected |
| Multi-Focus Municipality | - | Binary | 0.03 | 0.17 | LASSO |
| Core Municipality outside of Agglomeration | - | Binary | 0.03 | 0.16 | Not selected |
| Income | 4.7(lowest) - 6.9(highest) | Cardinal | 4.66 | 0.56 | Manually |
| Numer of Trips per Month and Station, Target Variable | Trips | Cardinal | 62.67 | 73.30 | - |

**Table 5**
All analyzed features.

# C. GWR and MGWR coefficient estimates

We detail the coefficients of our fitted models in Table 6 and Table 7. As expected, the vehicle supply and the population are the most important parameters. However, spatial information and socio-demographics play an important role and improve model performance.

| Feature | GWR coefficient values | | | | | | |
|---|---|---|---|---|---|---|---|
| | GWR Coefficient Mean | Standard Deviation | Min | Median | Max | Mean t-value $\pm\sigma$ | Significant Estimates [%] |
| Supply of Cars at a Station | 0.906 | 0.043 | 0.655 | 0.894 | 0.967 | 62.77 ± 7.50 | 100.00 |
| Population | 0.113 | 0.027 | 0.062 | 0.107 | 0.187 | 4.41 ± 1.18 | 94.86 |
| Public Transport Accessibility | 0.043 | 0.014 | 0.004 | 0.041 | 0.085 | 2.72 ± 0.86 | 59.62 |
| Proportion of Swiss Born Inhabitants | 0.037 | 0.015 | -0.017 | 0.04 | 0.077 | 1.64 ± 0.63 | 1.25 |
| Shopping POI Density | 0.036 | 0.019 | -0.021 | 0.041 | 0.07 | 2.19 ± 1.0 | 41.56 |
| Average Distance to Car-Sharing Station | 0.032 | 0.013 | -0.022 | 0.033 | 0.06 | 1.26 ± 0.49 | 0.00 |
| Number of Jobs | 0.023 | 0.025 | -0.024 | 0.031 | 0.094 | 1.14 ± 1.15 | 0.49 |
| Public POI Density | 0.021 | 0.02 | -0.027 | 0.02 | 0.06 | 1.17 ± 0.99 | 7.16 |
| Average Age | 0.019 | 0.015 | -0.065 | 0.016 | 0.059 | 0.98 ± 0.67 | 0.28 |
| Average Household Size | 0.015 | 0.012 | -0.02 | 0.015 | 0.049 | 0.57 ± 0.38 | 0.00 |
| Intercept | 0.006 | 0.013 | -0.029 | 0.004 | 0.041 | 0.45 ± 0.96 | 0.00 |
| Income | 0.003 | 0.029 | -0.029 | -0.011 | 0.057 | 0.005 ± 1.74 | 10.77 |
| Proportion of Female Inhabitants | 0.002 | 0.023 | -0.092 | 0.009 | 0.02 | 0.22 ± 0.91 | 5.49 |
| Distance to Restaurant | 0.001 | 0.028 | -0.081 | 0.014 | 0.027 | 0.07 ± 1.03 | 0.00 |
| Travel Time to Agglomerations by Car | 0.001 | 0.015 | -0.045 | 0.004 | 0.035 | 0.01± 0.49 | 0.00 |
| Cars per Household | -0.002 | 0.034 | -0.093 | 0.011 | 0.079 | 1.04 ± 0.10 | 0.14 |
| Distance to Small Shop | -0.003 | 0.023 | -0.027 | -0.013 | 0.071 | -0.42 ± 0.96 | 0.00 |
| Bicycles per Household | -0.005 | 0.015 | -0.057 | 0.003 | 0.013 | -0.20 ± 0.73 | 0.00 |
| Distance to Hospital | -0.005 | 0.013 | -0.058 | -0.003 | 0.026 | -0.22 ± 0.41 | 0.00 |
| Distance to Museum | -0.006 | 0.019 | -0.024 | -0.01 | 0.075 | -0.42 ± 0.68 | 0.00 |
| Distance to Postal Office | -0.006 | 0.016 | -0.031 | -0.01 | 0.039 | -0.48 ± 0.80 | 0.00 |
| High Income Municipalities (Binary) | -0.01 | 0.006 | -0.045 | -0.01 | -0.001 | -0.71 ± 0.31 | 0.00 |
| Multi-Focus Municipality (Binary) | -0.011 | 0.016 | -0.069 | -0.003 | 0 | -0.68 ± 0.72 | 5.49 |
| Number of Workplaces Sector 1 (Agriculture, Mining) | -0.011 | 0.013 | -0.032 | -0.017 | 0.017 | -0.75 ± 0.78 | 0.00 |
| Distance to Public Transport Station | -0.011 | 0.011 | -0.048 | -0.008 | -0.001 | 0.61 ± 0.41 | 0.00 |
| Rural Commuter Town (Binary) | -0.015 | 0.008 | -0.026 | -0.017 | 0.003 | -1.36 ± 0.72 | 0.00 |
| Station's Voronoi Area | -0.017 | 0.014 | -0.051 | -0.018 | 0.024 | -0.75 ± 0.61 | 0.00 |
| Travel Time to Agglomerations by Public Transport | -0.023 | 0.013 | -0.093 | -0.021 | 0.008 | -0.81 ± 0.42 | 0.28 |
| Distance to Doctor | -0.024 | 0.005 | -0.035 | -0.023 | 0.017 | -0.93 ± 0.21 | 0.00 |
| Motorbikes per Household | -0.025 | 0.007 | -0.044 | -0.023 | -0.01 | -1.76 ± 0.46 | 8.48 |
| Distance to Large Supermarket | -0.034 | 0.019 | -0.073 | -0.032 | 0.005 | -1.88 ± 1.15 | 29.33 |
| Accommodation POI Density | -0.052 | 0.01 | -0.085 | -0.049 | -0.042 | -2.98 ± 0.72 | 80.19 |
| Number of Competing Cars | -0.054 | 0.017 | -0.082 | -0.055 | -0.013 | -2.44 ± 0.83 | 57.19 |

**Table 6**
GWR coefficients upon model fitting.

| Feature | MGWR coefficient values | | | | | | | | | | |
|---|---|---|---|---|---|---|---|---|---|---|---|
| | Mean | Standard Deviation | Min | Median | Max | BW* (Neighbours) | BW* Min | BW* Median | BW* Max | Mean t-value $\pm\sigma$ | Significant Estimates [%] |
| Supply of Cars at a Station | 0.9075 | 0.1267 | 0.5524 | 0.9033 | 1.1726 | 232 | 6.4 km | 35.2 km | 156.8 km | 30.75 ± 9.55 | 100.00 |
| Population | 0.0964 | 0.0275 | 0.0665 | 0.0885 | 0.1641 | 864 | 55.6 km | 87.2 km | 217.6 km | 4.07±1.57 | 94.37 |
| Public Transport Accessibility | 0.0473 | 0.0011 | 0.0451 | 0.0470 | 0.0493 | 1438 | 157.7 km | 227.9 km | 330.7 km | 4.12± 0.12 | 100.00 |
| Intercept | 0.0360 | 0.0246 | -0.0168 | 0.0308 | 0.1076 | 623 | 36.2 km | 69.2 km | 192.8 km | 1.88± 1.19 | 15.15 |
| Number of Workplaces Sector 1 (Agriculture, Mining) | 0.0332 | 0.0845 | -0.1438 | 0.0241 | 0.3079 | 108 | 2.5 km | 18.4 km | 145.5 km | 0.45± 1.27 | 1.60 |
| Average Distance to Car-Sharing Station | 0.0326 | 0.0005 | 0.0311 | 0.0325 | 0.0333 | 1436 | 153.1 km | 226.8 km | 330.0 km | 1.84± 0.04 | 0.00 |
| Shopping POI Density | 0.0282 | 0.0029 | 0.0243 | 0.0268 | 0.0330 | 1438 | 157.7 km | 227.9 km | 330.7 km | 2.27± 0.24 | 65.18 |
| Income | 0.0250 | 0.0225 | 0.0002 | 0.0092 | 0.0602 | 750 | 50.4 km | 76.2 km | 197.6 km | 1.38± 1.21 | 23.63 |
| Proportion Swiss Born Inhabitants | 0.0210 | 0.0012 | 0.0114 | 0.0212 | 0.0228 | 1438 | 157.7 km | 227.9 km | 330.7 km | 1.24± 0.07 | 0.00 |
| Number of Jobs | 0.0200 | 0.0014 | 0.0153 | 0.0203 | 0.0226 | 1436 | 153.1 km | 226.8 km | 330.0 km | 1.25± 0.07 | 0.00 |
| Average Age | 0.0104 | 0.0020 | -0.0040 | 0.0107 | 0.0133 | 1438 | 157.7 km | 227.9 km | 330.7 km | 0.75± 0.14 | 0.00 |
| Travel Time to Agglomerations by Car | 0.0098 | 0.0050 | -0.0036 | 0.0077 | 0.0171 | 1432 | 151.0 km | 225.0 km | 329.2 km | 0.45± 0.22 | 0.00 |
| Proportion Female Inhabitants | 0.0083 | 0.0125 | -0.0265 | 0.0144 | 0.0168 | 1265 | 117.0 km | 169.9 km | 280.5 km | 0.57± 0.84 | 0.00 |
| Average Household Size | 0.0082 | 0.0009 | 0.0072 | 0.0078 | 0.0123 | 1438 | 157.7 km | 227.9 km | 330.7 km | 0.45± 0.06 | 0.00 |
| Public POI Density | 0.0072 | 0.0009 | 0.0050 | 0.0071 | 0.0098 | 1437 | 154.6 km | 227.6 km | 330.4 km | 0.53± 0.07 | 0.00 |
| Distance to Museum | 0.0031 | 0.0018 | 0.0009 | 0.0022 | 0.0078 | 1436 | 153.1 km | 226.8 km | 330.0 km | 0.22± 0.13 | 0.00 |
| Station's Voronoi Area | 0.0031 | 0.0014 | 0.0006 | 0.0025 | 0.0066 | 1409 | 145.5 km | 220.8 km | 325.6 km | 0.19± 0.08 | 0.00 |
| Distance to Postal Office | -0.0003 | 0.0256 | -0.0732 | 0.0081 | 0.0308 | 498 | 28.0 km | 60.4 km | 174.9 km | -0.01± 1.16 | 8.48 |
| Distance to Hospital | -0.0040 | 0.0018 | -0.0058 | -0.0050 | 0.0023 | 1409 | 145.5 km | 220.8 km | 325.6 km | -0.28± 0.13 | 0.00 |
| Distance to Small Shop | -0.0077 | 0.0033 | -0.0114 | -0.0098 | -0.0004 | 1409 | 145.5 km | 220.8 km | 325.6 km | -0.61± 0.27 | 0.00 |
| High Income Municipalities (Binary) | -0.0080 | 0.0009 | -0.0103 | -0.0077 | -0.0057 | 1436 | 153.1 km | 226.8 km | 330.0 km | -0.78± 0.08 | 0.00 |
| Multi-Focus Municipality (Binary) | -0.0090 | 0.0064 | -0.0227 | -0.0052 | -0.0034 | 1331 | 127.4 km | 174.4 km | 286.6 km | -0.82± 0.49 | 0.00 |
| Distance to Public Transport Station | -0.0098 | 0.0013 | -0.0134 | -0.0091 | -0.0086 | 1436 | 153.1 km | 226.8 km | 330.0 km | -0.82± 0.1 | 0.00 |
| Motorbikes per Household | -0.0118 | 0.0057 | -0.0252 | -0.0085 | -0.0071 | 1361 | 141.1 km | 211.4 km | 317.8 km | -1.1± 0.48 | 0.00 |
| Distance to Restaurant | -0.0118 | 0.0062 | -0.0244 | -0.0083 | -0.0062 | 1391 | 143.7 km | 219.2 km | 324.6 km | -0.71± 0.37 | 0.00 |
| Bicycles per Household | -0.0125 | 0.0021 | -0.0176 | -0.0113 | -0.0103 | 1436 | 153.1 km | 226.8 km | 330.0 km | -0.99± 0.15 | 0.00 |
| Rural Commuter Town (Binary) | -0.0143 | 0.0048 | -0.0186 | -0.0171 | -0.0039 | 1265 | 117.0 km | 169.9 km | 280.5 km | -1.48± 0.51 | 0.00 |
| Cars per Household | -0.0198 | 0.0024 | -0.0250 | -0.0182 | -0.0176 | 1438 | 157.7 km | 227.9 km | 330.7 km | -1.07± 0.13 | 0.00 |
| Distance to Doctor | -0.0203 | 0.0012 | -0.0219 | -0.0208 | -0.0165 | 1438 | 157.7 km | 227.9 km | 330.7 km | -1.23± 0.09 | 0.00 |
| Number of Competing Cars | -0.0272 | 0.0030 | -0.0311 | -0.0290 | -0.0217 | 1438 | 157.7 km | 227.9 km | 330.7 km | -1.66± 0.13 | 0.00 |
| Accommodation POI Density | -0.0287 | 0.0036 | -0.0364 | -0.0266 | -0.0256 | 1409 | 145.5 km | 220.8 km | 325.6 km | -2.14± 0.32 | 34.89 |
| Distance to Large Supermarket | -0.0374 | 0.0007 | -0.0387 | -0.0375 | -0.0354 | 1438 | 157.7 km | 227.9 km | 330.7 km | -3.01± 0.2 | 100.00 |
| Travel Time to Agglomerations by Public Transport | -0.0380 | 0.0033 | -0.0518 | -0.0391 | -0.0330 | 1409 | 145.5 km | 220.8 km | 325.6 km | -1.74 ± 0.15 | 1.25 |

*BW = Bandwidth

**Table 7**
MGWR coefficients upon model fitting.